\documentclass[11pt,a4paper]{article}
\usepackage[hyperref]{naaclhlt2018}
\usepackage[english]{babel}
\usepackage{times}
\usepackage{latexsym}
\usepackage{subcaption}

\usepackage{CJKutf8}
\usepackage{array}
\usepackage{graphicx}
\usepackage{tabularx}
\usepackage{multirow}
\usepackage{algorithm}
\usepackage{algpseudocode}
\usepackage{amsmath}
\usepackage{tikz-dependency}
\usepackage[group-separator={,},group-minimum-digits={3}]{siunitx}

\newcommand{\chn}[1]{\begin{CJK*}{UTF8}{gbsn}#1\end{CJK*}}

\makeatletter
\newcommand{\StatexIndent}[1][3]{%
  \setlength\@tempdima{\algorithmicindent}%
  \Statex\hskip\dimexpr#1\@tempdima\relax}
\algdef{C}[IF]{IF}{ElsIfNoThen}[1]{\algorithmicelse\ \algorithmicif\ #1}%
\algdef{C}[IF]{IF}{IfNoThen}[1]{\algorithmicif\ #1}%
\makeatother

\aclfinalcopy 

\title{Hybrid Oracle: Making Use of Ambiguity in Transition-based Chinese Dependency Parsing}

\author{Xuancheng Ren \qquad Xu Sun\\
  MOE Key Laboratory of Computational Linguistics\\
  School of Electronics Engineering and Computer Science\\ Peking University\\ Beijing, China \\
  {\tt \{renxc,xusun\}@pku.edu.cn}
}
\date{}

\begin{document}
\maketitle
\begin{abstract}
In the training of transition-based dependency parsers, an oracle is used to predict a transition sequence for a sentence and its gold tree. However, the transition system may exhibit ambiguity, that is, there can be multiple correct transition sequences that form the gold tree. We propose to make use of the property in the training of neural dependency parsers, and present the \textit{Hybrid Oracle}. The new oracle gives all the correct transitions for a parsing state, which are used in the cross entropy loss function to provide better supervisory signal. It is also used to generate different transition sequences for a sentence to better explore the training data and improve the generalization ability of the parser. Evaluations show that the parsers trained using the hybrid oracle outperform the parsers using the traditional oracle in Chinese dependency parsing. We provide analysis from a linguistic view. \footnote{The code is available at \url{https://github.com/lancopku/nndep}.}
\end{abstract}

\section{Introduction}

Transition-based dependency parsers treat parsing as a sequence of actions that form a parse tree. The actions are called \textit{transitions}. At each step of parsing, a classifier is used to select the best transition, based on the parsing state of that step. The parsing state is called \textit{configuration}. In terms of the definition of the transitions and the configurations, common transition-based systems\footnote{In this paper, we focus on projective dependency parsing.} can be broadly categorized into arc-standard \citep{Nivre:2004:WIP,Nivre08}, arc-eager \citep{Nivre:2003:IWPT}, and arc-hybrid \citep{kuhlmann-gomezrodriguez-satta:2011:ACL-HLT2011} systems. 

To train such a parser, an optimal transition is needed for each configuration, and the mechanism to generate the optimal transition is called \textit{oracle}. Traditionally, parsers are only trained on configurations that result from the optimal transitions in a static way. However, the static oracles face two major issues. 

\begin{figure}[t]
    \centering
    \subcaptionbox{The dependency parse tree.}{
        \begin{dependency}
            \begin{deptext}[column sep=18pt]
            \chn{在} \& \chn{文} \& \chn{中} \\
            in \& text \& within \\
            ``in \& the \& text'' \\
            \end{deptext}
            \depedge{2}{1}{case}
            \depedge{2}{3}{case}
            \deproot[edge unit distance=1.5ex]{2}{root}
        \end{dependency}
        }
    \quad
    \subcaptionbox{The correct transition sequences.}{
        \begin{tabular}{c@{ }c}
        & \\
           (1) & shift shift larc shift rarc \\
           (2) & shift shift shift rarc larc
        \end{tabular}
    }
    \caption{An example of the dependency parse tree of a phrase that exposes the ambiguity of the arc-standard system. As we can see, for a constituent with dependents both on the left and on the right, multiple transition sequences can lead to the gold parse tree of the constituent.}
    \label{fig:dep}
\vspace{-1\baselineskip}
\end{figure}

First, due to the spurious ambiguity of the transition systems, there can be multiple correct transition sequences that could lead to the gold parse tree. An example is shown in Figure~\ref{fig:dep}. The static oracles deal with the issue by setting preference. For example, when $shift/larc$ ambiguity happens in arc-standard systems, the oracle gives $larc$ as the optimal, e.g. the sequence (1) in the example. 

Second, at prediction time, wrong transition decisions will lead to the configurations that cannot lead to the gold parse tree. As the parser may never be trained on such type of configurations, error at this step may propagate to the following steps. \citet{TACL145} propose dynamic oracles that define the optimal transition for any configuration, whether the configuration can lead to the gold parse tree or not. They show that the computation is straight-forward for arc-eager and arc-hybrid systems, which satisfy arc-decomposability. However, such property does not hold for arc-standard systems. Although dynamic oracles for arc-standard systems can be computed using a dynamic programming algorithm which is polynomial in the length of the stack~\citep{goldberg-etal:2014:tacl}, an efficient dynamic oracle for arc-standard systems remains undefined.

In this work, we define \textit{Hybrid Oracle} for arc-standard systems, which identifies all the feasible transitions when ambiguity happens, and can be used to explore all the correct configurations during training. The oracle can be implemented in linear time to the sentence length. Experiments show that by using the hybrid oracle, the performance of the parser can be improved compared to the baseline for Chinese dependency parsing on CTB5 and CoNLL17. We give analysis on the reason why the hybrid oracle is useful for Chinese dependency parsing from a syntactic view.

\section{Proposed Method}

With the application of neural classifiers, we are provided with new insights of tackling the issues faced by the static oracles. We first introduce the arc-standard system, and the commonly used static oracle. Then, we define hybrid oracle for the arc-standard system, and show how the correct transitions can be incorporated into the loss function of the neural classifiers, and how the hybrid oracle can be used to explore the correct configurations.

\textbf{The Arc-Standard System} Many transition systems exist in the literature. Hybrid oracle can be defined for transition-based systems that bear ambiguity, e.g. arc-eager systems. In this work, we focus on the arc-standard system, which is more popular for neural networks, and often produces the best results, even for which an efficient dynamic oracle is not available. In the arc-standard system, a configuration $c=(\sigma, \beta, A)$ consists of a stack $\sigma$, a buffer $\beta$, and a set $A$ of dependency arcs. Given a sentence $S=w_1,w_2,\ldots,w_n$, the system is initialized with $\sigma=\phi$, $\beta=ROOT,w_1,w_2,\ldots,w_n$, and $A=\phi$, where $ROOT$ represents the head of the root word. The terminal configuration is $\sigma=ROOT$, $\beta=\phi$, and $|A|=n$. There are three types of transitions in the system, $larc_l$, $rarc_l$, and $shift$ defined as:
\begin{equation*}
\small
\begin{split}
    shift[(\sigma, b|\beta, A)] & = (\sigma|b, \beta, A) \\
    rarc_l[(\sigma|s_1|s_0, \beta, A)] & = (\sigma|s_1, \beta, A\cup\{\langle s_1, l, s_0 \rangle\}) \\
    larc_l[(\sigma|s_1|s_0, \beta, A)] & = (\sigma|s_0, \beta, A\cup\{\langle s_0, l, s_1 \rangle\}) 
\end{split}
\end{equation*}

The ambiguity happens in the arc-standard system when the constituent at the stack top has unattached left dependents and unattached right dependents. Under the circumstance, both $shift$ and $larc_l$ are correct transitions, but will lead to different configurations.

\textbf{Standard Oracle}
The oracle commonly used in the literature, which we name \textit{Standard Oracle}, selects the $larc_l$ as the correct transition when ambiguity happens, which follows the intuition that the dependents should be attached as early as possible, i.e., the left dependents should be attached before the right dependents. This preference also makes the stack of the configuration shortest. The intuition is also shared by arc-eager and arc-hybrid systems. This strategy works well for English, and has been apply to many other language.

\textbf{Hybrid Oracle}
The main difference between the static oracle and the hybrid oracle is that the hybrid oracle identifies all the correct transitions for a given configuration. For arc-standard systems, the implementation is straight-forward. When $larc$ is feasible for a configuration, the algorithm further checks if the constituent at the stack top has unattached right dependents. If so, $shift$ is also feasible. 

\begin{table*}[t]
\centering
\setlength{\tabcolsep}{2pt}
\begin{tabular}{l| r r| r r| r r r}
\hline
CTB5 & \#sentences & \#tokens & \#left dep. & \#right dep. & \#amb. sentences& \#amb. heads & \#amb. tokens  \\
 \hline
Train & \num{16091} & \num{437991}  & \num{288507} & \num{133393} &\num{14935} & \num{48040} & \num{215799} \\
Dev. & \num{803} & \num{20454}  & \num{13222} & \num{6429} & \num{746} & \num{2277} &\num{10071} \\
Test & \num{1910} & \num{50319}  & \num{33231} & \num{15178} & \num{1772} & \num{5588} &\num{24948} \\
\hline
\hline
CoNLL17 & \#sentences & \#tokens & \#left dep. & \#right dep. & \#amb. sentences& \#amb. heads & \#amb. tokens  \\
 \hline
Train & \num{3997} & \num{98608}  & \num{67985} & \num{26626} &\num{3993} & \num{12967} & \num{54996} \\
Dev. & \num{500} & \num{12663}  & \num{8799} & \num{3364} & \num{500} & \num{1676} &\num{7160} \\
Test & \num{500} & \num{12012}  & \num{8268} & \num{3244} & \num{500} & \num{1530} &\num{6594} \\
\hline
\end{tabular}
\caption{Statistics of the datasets. \textit{\#left dep.} is the number of the constituents that are on the left of their heads. \textit{\#right dep.} is defined similarly. \textit{\#amb. sentences} is the number of the sentences that have multiple correct transition sequences. \textit{\#amb. heads} is the number of the constituents that cause the ambiguity, i.e., the constituents have both left dependents and right dependents.
}
\label{tab:datasets}
\vspace{-0.75\baselineskip}
\end{table*}

The key part is that this information can be learned by the neural network through the loss function. Formally, the widely-used cross entropy loss function is defined as:
\begin{equation*}
    J(\pmb{x}, \pmb{y}; \theta) = -\sum_{i=0}^{m} \pmb{y}_i \log f(\pmb{x})_i
\end{equation*}
where $\pmb{x}$ is the input vector, $\pmb{y}$ is the target vector, $m$ is the number of labels, and $f$ stands for the neural network. Most of the time, the target vector $\pmb{y}$ is the one-hot representation of the true label, making the loss function the same as the negative log likelihood loss. However, the target vector $\pmb{y}$ is essentially the probability distribution of the labels. When there are multiple true labels, a simple strategy is to make the true labels uniformly distributed, i.e., if there are $l$ true labels, when $i$ is the index of a true label, we could make $\pmb{y}_i = l^{-1}$, which then forms the target probability distribution of the labels. The correct transitions given by the hybrid oracle act as the multiple true labels, and in this way, the neural network is provided refined supervisory signals, and will better conform to the data and the system.

Moreover, the hybrid oracle can be used to explore the correct transition sequences during the training. As the training example for the neural network is the configuration and its transitions, the examples can generated on-the-fly in training step-by-step. The transition applied at each step can be selected randomly from the correct transitions, so that the transition sequence for a sentence would be different at different epoch, which leads to more training data and may improve the generalization ability of the neural network.

\begin{algorithm}[H]
    \footnotesize
	\caption{Hybrid Oracle for Arc-Standard \label{alg}}
	\begin{algorithmic}[1]
		\Procedure{HybridOracle}{$c, A_{gold}$}
			\If{$c = (\sigma|s_1|s_0, \beta, A)$ \textbf{ and } $\langle s_0, l, s_1\rangle \in A_{gold} $}
			    \If{$\Call{HasRightUnattached}{s_0}$}
			        \If {$\Call{Uniform}{0, 1}<0.5$}
			            \State $t \gets larc_{label}$
			        \Else 
			            \State $t \gets shift$
			        \EndIf
			        \State $ct \gets [shift, larc_{l}]$
			    \Else
			        \State $t \gets larc_{label}$
    			    \State $ct \gets [larc_{l}]$
    			\EndIf
			\ElsIfNoThen {$c = (\sigma|s_1|s_0, \beta, A) \textbf{ and } \langle s_1, l, s_0 \rangle \in A_{gold}$}
			 \StatexIndent $\textbf{and not}\ \Call{HasUnattached}{s_0} \ \algorithmicthen$
			    \State $t \gets rarc_{l}$
			    \State $ct \gets [rarc_{l}]$
			\Else
			   \State $t \gets shift$
			   \State $ct \gets [shift]$
			\EndIf
			\State \Return $t, ct$
		\EndProcedure
	\end{algorithmic}
\end{algorithm}

The combined procedure is shown in Algorithm~\ref{alg}, which gives the correct transitions $c_t$ for a configuration, and the random transition $t$ that could be applied to generate the next configuration. The oracle can be implemented in $O(n)$ time, where $n$ is the length of the sentence, because of the checking of the unattached dependents. If preprocessing is done and the dependents of a constituent are recorded, the checking can be done in $O(1)$ time. Therefore, the oracle can be implemented in constant time.

\section{Experiments}

We compare the standard oracle and the hybrid oracle using neural models on Chinese Treebank 5.0 (CTB5)~\citep{Xue:2005:PCT:1064781.1064785} and the Chinese treebank from CoNLL 2017 shared task (CoNLL17)~\citep{ud20data,ud20testdata}. For CTB5, the phrase-structure treebank is converted into dependency annotations as described in \citet{chen-manning:2014:EMNLP2014}. We use standard training, validation, and test splits. The statistics are listed in Table~\ref{tab:datasets}. The evaluation metrics are unlabeled attachment score (UAS), labeled attachment score (LAS), and unlabeled exact match (UEM). Following previous work, we do not include the punctuations in the evaluation\footnote{A token is a punctuation if its gold POS tag is \{`` '' : , .\} for CoNLL17 and PU for Chinese.}.

Following \citet{cross-huang:2016:ACL,TACL885}, we use bidirectional LSTM~\citep{HochreiterS97} to extract the context features for a constituent, and use two feed-forward networks to classify the transition (i.e., $shift$, $larc$, and $rarc$) and the label, respectively. The input features of the feed-forward networks are the same and include the context features of the two topmost element at the stack and the first element in the buffer. Following previous work, we use gold POS tags in the training\footnote{In particular, we use language-specific POS tags for CoNLL17.}.

Because the proposed method is irrelevant of the classifier type, we conduct minimal hyper-parameter tuning in the experiments. We do not use pre-trained word embeddings, and the embedding dimensions for words and POS tags are set the same to 50. The BLSTM network has one layer and contains 200 dimension at each direction. The hidden dimension of the feed-forward network is also set to 200. We apply dropout~\citep{HintonSKSS12} with 0.5 for the output of BLSTM. The optimizer used is Adam~\citep{KingmaB14} with default settings. The mini-batch is 1 sentence, and we trained 20 epochs, and the epoch achieving the best UAS score is chosen for evaluation.

\subsection{Experimental Results}

\begin{table}[ht]
\centering
\setlength{\tabcolsep}{2pt}
\begin{tabular}{l | c c c |c c c }
\hline
\multirow{2}{*}{method} & \multicolumn{3}{c|}{CTB5.0 Dev.} & \multicolumn{3}{c}{CTB5.0 Test}  \\
  & UAS & LAS & UEM & UAS & LAS & UEM  \\
\hline
standard & 85.00 & 83.19 & 32.63 & 85.03 & 83.40 & 31.20  \\
hybrid & \bf 85.79 & \bf 84.06 & \bf 33.30 & \bf 85.68 & \bf 83.90 & \bf 33.50 \\
\hline
\end{tabular}
\caption{Results on CTB5. As we can see, the parser trained with the hybrid oracle consistently outperforms the one with standard oracle.}
\label{tab:res-ctb}
\vspace{-0.75\baselineskip}
\end{table}

\begin{table}[ht]
\centering
\setlength{\tabcolsep}{2pt}
\begin{tabular}{l | c c c |c c c }
\hline
\multirow{2}{*}{method} & \multicolumn{3}{c|}{CoNLL17 Dev.} & \multicolumn{3}{c}{CoNLL17 Test.} \\
  & UAS & LAS & UEM & UAS & LAS & UEM \\
\hline
standard &  81.38 & 77.68 & 22.00 & 82.82 & 79.27 & 25.40  \\
hybrid  & \bf 81.91 & \bf 78.14 & \bf 22.60 & \bf 83.65 & \bf 80.19 & \bf 27.20\\
\hline
\end{tabular}
\caption{Results on CoNLL17. The hybrid oracle improves the dependency parsing on the dataset as well, in terms of the metrics.}
\label{tab:res-conll}
\end{table}

The results on CTB5 and CoNLL17 are summarized in Table~\ref{tab:res-ctb} and Table~\ref{tab:res-conll}, respectively. As we can see, for the Chinese treebanks, the parsers trained using Hybrid Oracle perform substantially well compared to the Standard Oracle, in terms of all the metrics. For UAS, the improvements on the test sets of CTB5 and CoNLL17 are 0.65 and 0.83, respectively. For LAS, the improvements on the test sets of CTB5 and CoNLL17 are 0.50 and 0.92. Hybrid oracle improves the UEM metric by 2.30 on the test set of CTB5 and by 1.80 on the test set of CoNLL17. The results on the development sets are similar.

\begin{figure}[ht]
    \centering
    \includegraphics[width=0.6\linewidth]{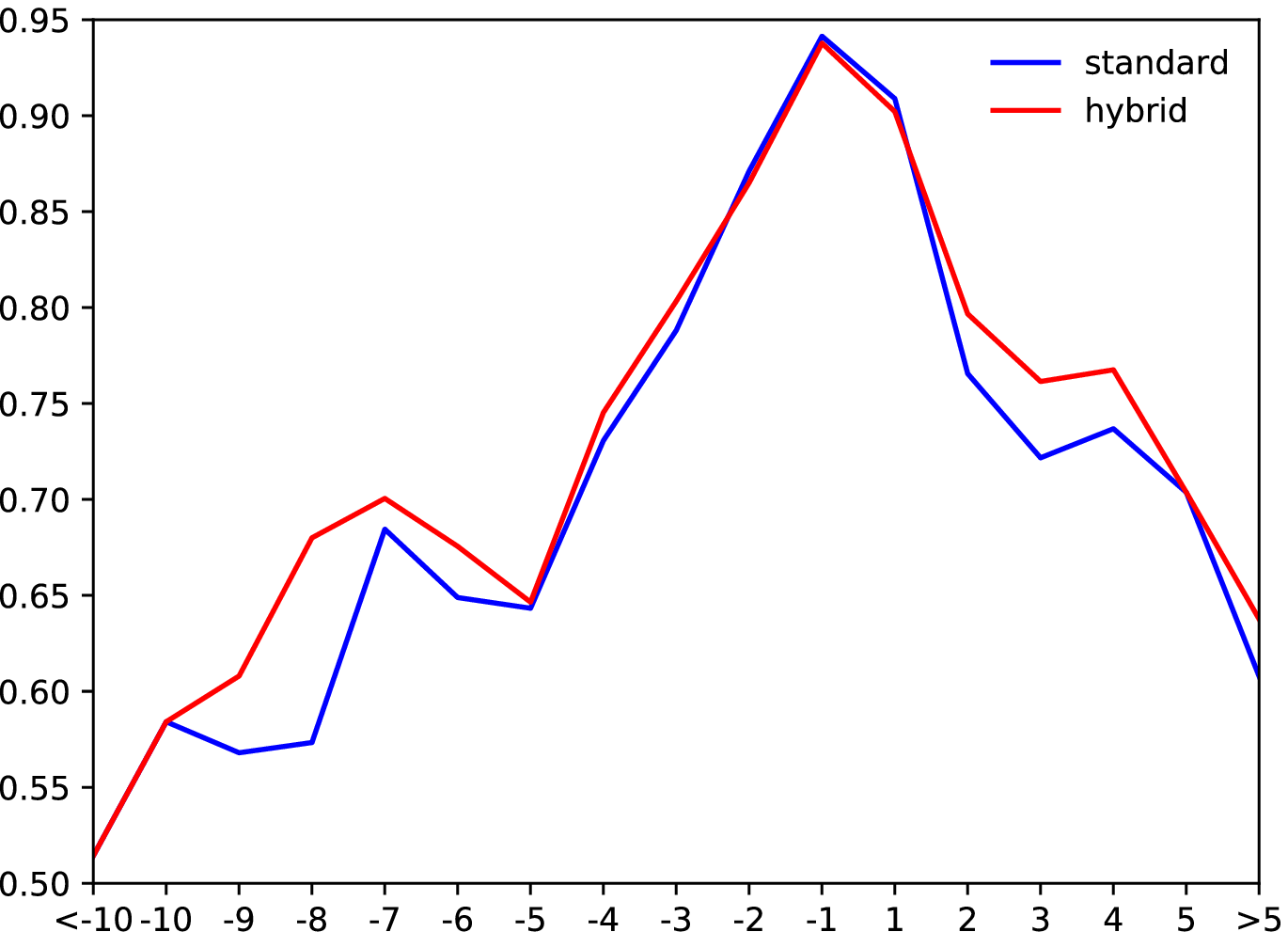}
    \caption{Recall regarding to length of dependency arcs of in development set of CoNLL17. If the length is negative, it means the dependents are on the left of their heads. For the lengths of which there are less than 100 arcs, they are aggregated. We can see that dependents on the left are dominant. The hybrid oracle is helpful when predicting long arcs. \label{fig:arc}}
\vspace{-0.75\baselineskip}
\end{figure}

To see why the hybrid oracle is helpful to the parsing, we further examine the effect of using the hybrid oracle on the recall of the dependency arcs. Figure~\ref{fig:arc} shows the results on CoNLL17. We can see that when parsing longer arcs, using the hybrid oracle is more helpful. When the dependent is on the left, the improvement is better.

We believe the phenomenon may be explained from the linguistic view of branching. In linguistics, branching refers to the shape of the parse trees that represent the structure of sentences. Chinese is a principally left-branching language in that relative clauses and sub-ordinate clauses position to be the left of their head. 
As shown in \citet{liu2010dependency}, in Chinese dependency treebanks, left arcs are dominant and double the right arcs. The statistics of the datasets we used also support the claim.

The property suggests that the parse of the left of a constituent in Chinese is more difficult than the parse of the right of a constituent. It may not be wise to eagerly attach the left dependents to their head. Training using the standard oracle, the parser is forced to always attach the left dependents first, and always forms the shortest stack. It may be more prone to errors, as the information is less complete when parsing the left branch. When training using the hybrid oracle, if larc and shift are both feasible, it is up to the parser to determine which transition to apply according to its confidence of the transition. 

The improvements on CoNLL17 are better most of the time. This could be also explained from the properties of the data. For CTB5, left dependents account for 67.28\% of the development set, while for CoNLL 17, left dependents account for 72.24\% of the development set. 49.81\% tokens in the development set of CTB5 can be parsed by multiple sequences, while the ratio of the development set of CoNLL17 is 56.54\%. These properties suggest that for CoNLL17, there is more ambiguity that can be exploited.

\section{Related Work}

For Chinese dependency parsing, many transition-based methods using neural networks have been proposed. For CTB5, the best UAS achieved with models other than the baseline in the similar setting is 85.7 \citep{dyer-EtAl:2015:ACL-IJCNLP}\footnote{Unfortunately, we cannot find the results to two decimal places. As shown in \citet{ballesteros-EtAl:2016:emnlp}, a rerun gets a UAS of 85.48.}. The proposed method achieves a UAS of 85.68. As reported in \citet{cross-huang:2016:ACL}, when using a two-layer BLSTM, the baseline model can reach UAS of 86.35. We believe our result can be further improved if using the same technique. 
In other settings, where external data are allowed and the efficiency of the parsing can be compromised, using techniques, such as beam search, pre-trained word embeddings, or refined feature engineering, the UAS on CTB5 can be improved to 87.6~\citep{zhang-nivre:2011:acl,ballesteros-EtAl:2016:emnlp,TACL885}. 
For the parsing of the Chinese Treebank in CoNLL 2017 shared task, as the dataset is relatively new, we cannot find sufficient results in the similar setting to compare.

Other related work includes \citet{sartorio-etal:2013:acl} and \citet{ballesteros-EtAl:2016:emnlp}. \citet{sartorio-etal:2013:acl} propose to guide the model using the transition with the maximum score assigned by the model, when there are multiple correct transitions in their proposed transition system. They call the strategy easy-first. \citet{ballesteros-EtAl:2016:emnlp} propose to maximize the marginal likelihood of all correct transitions in arc-hybrid systems. The fact is that although there can be multiple correct transitions, the training procedure tie-breaks according to the model's current prediction in their methods. The model still learns only one transition for a configuration, and the oracle path becomes stable with the training. In our method, the model learns all the correct transitions, with no preference either manually set or learned. We believe this is more beneficial for neural models and leads to better generalization ability, as the neural models tend to over-fit the data as shown by~\citet{ballesteros-EtAl:2016:emnlp,TACL885}.

\section{Conclusion and Future Work}

We have presented a new oracle for training arc-standard transition-based dependency parsers. The new oracle gives all of the correct transitions for a configuration, which can be used for training the classifier with better supervisory signal. To make use of the multiple correct transitions, in training, a random correct transition is chosen when conduct transitions, which has the effect of increasing the training data. Experiments show that our method achieves better performance compared with the standard oracles in Chinese dependency parsing on two datasets. The analysis shows that the hybrid oracle may also works for other language that is left-branching. We would like to conduct experiments on other language to see whether the claim holds.

\section*{Acknowledgments}

This work was supported in part by National Natural Science Foundation of China (No. 61673028), and an Okawa Research Grant (2016).

\nocite{SunM17,SunRMW17,SunRMWLW17}


\bibliography{naaclhlt2018}
\bibliographystyle{acl_natbib}


\end{document}